\documentclass[technote,a4paper,leqno]{IEEEtran}
\pdfoutput=1

\usepackage[utf8]{inputenc}
\usepackage[T1]{fontenc}
\usepackage{amsmath,amssymb}
\usepackage[pdftex]{graphicx}
\usepackage{caption}
\usepackage[hang]{subfigure}
\usepackage{url}
\usepackage[super]{nth}
\usepackage[breaklinks]{hyperref}

\usepackage[absolute,overlay]{textpos}
\usepackage{tikz}
\usepackage{csquotes}
\usepackage{booktabs}  % for \toprule, \midrule and \bottomrule
\usepackage{pbox}
\usepackage{braket} % needed for nice printing of sets
\usepackage[english]{babel}
\usepackage{mathtools}
\usepackage{xspace}
\usepackage[inner=2.5cm,outer=2.5cm,top=2.5cm,bottom=2cm,includeheadfoot]{geometry}
\usepackage{glossaries}
\newacronym{ACM}{ACM}{active contour model}
\newacronym{BOV}{BOV}{bag-of-visual-words}
\newacronym{CRF}{CRF}{Conditional Random Field}
\newacronym{GPU}{GPU}{graphics processing unit}
\newacronym{HOG}{HOG}{histogram of oriented gradients}
\newacronym{HSV}{HSV}{hue saturation value}
\newacronym{ILSVRC}{ILSVRC}{ImageNet Large-Scale Visual Recognition Challenge}
\newacronym{ICM}{ICM}{Iterated Conditional Mode}
\newacronym{MAP}{MAP}{Maximum A Posteriori}
\newacronym{MLE}{MLE}{maximum likelihood estimation}
\newacronym{MR}{MR}{magnetic resonance}
\newacronym{MRF}{MRF}{Markov Random Field}
\newacronym{PCA}{PCA}{principal component analysis}
\newacronym{PGM}{PGM}{probabilistic graphical model}
\newacronym{RBF}{RBF}{radial basis function}
\newacronym{STF}{STF}{Semantic Texton Forest}
\newacronym{SIFT}{SIFT}{scale-invariant feature transform}
\newacronym{CNN}{CNN}{Convolution Neuronal Network}
\newacronym{MLP}{MLP}{Multilayer perceptron}
\newacronym{SVM}{SVM}{Support Vector Machine}

\makeglossaries

\usepackage[nameinlink, noabbrev,capitalise]{cleveref} % has to be after hyperref, nthe l.137 ...eam attr{/N 4}  file{sRGBIEC1966-2.1.icmorem, amsthm

\usepackage{enumitem}
\crefname{problemnr}{\textup{problem}}{\textup{problems}}
\newlist{problemnr}{enumerate}{1}
\setlist[problemnr]{label={P\arabic*}, ref=P\arabic*}
\crefalias{problemnri}{problemnr}

\newcommand*{\eg}{e.g.\@\xspace}

\usepackage[binary-units,group-separator={,}]{siunitx}
\DeclareSIUnit\pixel{px}
\usepackage{microtype}

\DeclareGraphicsExtensions{.jpg}
\DeclareMathOperator{\sgn}{sgn}
\DeclareMathOperator*{\minimize}{minimize}
\DeclareMathOperator*{\maximize}{maximize}

\newcommand{\titleboth}{A Survey of Semantic Segmentation}
\title{\titleboth}
\author{Martin Thoma\\% ORCID: http://orcid.org/0000-0002-6517-1690 Thoma, Martin
info@martin-thoma.de
}

\hypersetup{
    pdfauthor   = {Martin Thoma},
    pdfkeywords = {semantic segmentation, pixelwise classification, pixel-level classification},
    pdfsubject  = {Segmentation},
    pdftitle    = {\titleboth},
}

\begin{document}
\maketitle
%!TEX root = vorlage.tex

\begin{abstract}
This survey gives an overview over different techniques used for pixel-level
semantic segmentation. Metrics and datasets for the evaluation of segmentation
algorithms and traditional approaches for segmentation such as unsupervised
methods, Decision Forests and SVMs are described and pointers to the relevant
papers are given. Recently published approaches with convolutional neural
networks are mentioned and typical problematic situations for segmentation
algorithms are examined. A taxonomy of segmentation algorithms is given.
\end{abstract}

%!TEX root = vorlage.tex

\section{Introduction}\label{sec:introduction}
Semantic segmentation is the task of clustering parts of images together which
belong to the same object class. This type of algorithm has several use-cases
such as detecting road signs~\cite{4220659}, detecting
tumors~\cite{moon2002automatic}, detecting medical instruments in
operations~\cite{wei1997automatic}, colon crypts
segmentation~\cite{cohen2015memory}, land use and land cover
classification~\cite{huang2002assessment}. In contrast, non-semantic
segmentation only clusters pixels together based on general characteristics of
single objects. Hence the task of non-semantic segmentation is not
well-defined, as many different segmentations might be acceptable.

Several applications of segmentation in medicine are listed
in~\cite{annurev.bioeng.2.1.315}.

Object detection, in comparison to semantic segmentation, has to distinguish
different instances of the same object. While having a semantic segmentation is
certainly a big advantage when trying to get object instances, there are a
couple of problems: neighboring pixels of the same class might belong to
different object instances and regions which are not connected my belong to the
same object instance. For example, a tree in front of a car which visually
divides the car into two parts.

This paper is organized as follows: It begins by giving a taxonomy of
segmentation algorithms in \cref{sec:taxonomy}. A summary of quality measures
and datasets which are used for semantic segmentation follows in
\cref{sec:evaluation-and-datasets}. A summary of traditional
segmentation algorithms and their characteristics follows in
\cref{sec:traditional-approaches}, as well as a brief, non-exhaustive
summary of recently published semantic segmentation algorithms which are based
on neural networks in \cref{sec:nn}. Finally, \cref{sec:problems} informs the
reader about typical problematic cases for segmentation algorithms.

%!TEX root = vorlage.tex

\section{Taxonomy of Segmentation Algorithms}\label{sec:taxonomy}
The computer vision community has published a wide range of segmentation
algorithms so far. Those algorithms can be grouped by the kind of data they
operate on and the kind of segmentation they are able to produce.

The following subsections will give four different criteria by which
segmentation algorithms can be classified.

This survey describes fixed-class (see \cref{subsec:allowed-classes}),
single-class affiliation (see \cref{subsec:class-affiliation}) algorithms which
work on grayscale or colored single pixel images (see \cref{subsec:input-data})
in a completely automated, passive fashion (see \cref{subsec:operation-state}).

\subsection{Allowed classes}\label{subsec:allowed-classes}
Semantic segmentation is a classification task. As such, the classes on which
the algorithm is trained is a central design decision.

Most algorithms work with a fixed set of classes; some even only work on binary
classes like \textit{foreground vs
background}~\cite{4228537,carreira2010constrained} or \textit{street vs no
street}~\cite{bittel2015pixel}.

However, there are also unsupervised segmentation algorithms which do not
distinguish classes at all (see
\cref{subsec:unsupervised-traditional-segmentation}) as well as segmentation
algorithms which are able to recognize when they don't know a class. For
example, in~\cite{gould2008multi} a
\textbf{void class} was added for classes which were not in the training set.
Such a void class was also used in the MSRCv2 dataset (see
\cref{subsubsec:MSRCv2}) to make it possible to make more coarse segmentations
and thus having to spend less time annotating the image.

\subsection{Class affiliation of pixels}\label{subsec:class-affiliation}
Humans do an incredible job when looking at the world. For example, when we see
a glass of water standing on a table we can automatically say that there is the
glass and behind it the table, even if we only had a single image and were not
allowed to move. This means we simultaneously two labels to the coordinates of
the glass: Glass and table. Although there is much more work being done on
\textbf{single class affiliation} segmentation algorithms, there is a
publication about \textbf{multiple class affiliation}
segmentation~\cite{levin2008spectral}. Similarly, recent publications in
pixel-level object segmentation used layered models~\cite{yang2012layered}.

\goodbreak
\subsection{Input Data}\label{subsec:input-data}
The available data which can be used for the inference of a segmentation varies
by application.

\begin{itemize}
    \item \textbf{Grayscale vs colored}: Grayscale images are commonly used in
          medical imaging such as \gls{MR} imaging or ultrasonography whereas
          colored photographs are obviously widespread.
    \item \textbf{Excluding or including depth data}: RGB-D, sometimes also
          called range~\cite{hoover1996experimental} is available in robotics,
          autonomous cars and recently also in consumer electronics such as
          Microsoft Kinect~\cite{6190806}.
    \item \textbf{Single image vs stereo images vs co-segmentation}: Single
          image segmentation is the most wide-spread kind of segmentation, but
          using stereo images was already tried in~\cite{boykov2001fast}. It
          can be seen as a more natural way of segmentation as most mammals
          have two eyes. It can also be seen as being related to having
          depth data.\\
          Co-segmentation as in~\cite{1640859,collins2012random} is the problem
          of finding a consistent segmentation for multiple images. This problem
          can be seen in two ways: One the one hand, it can be seen as the problem
          of finding common objects in at least two images. On the other hand,
          every image after the first can be used as an additional source of
          information to find a meaningful segmentation. This idea can be
          extended to time series such as videos.
    \item \textbf{2D vs 3D}: Segmenting images is a 2D~segmentation task where
          the smallest unit is called a \textit{pixel}. In 3D data, such as
          volumetric X-ray CT images as they were used in~\cite{929615}, the
          smallest unit is called a voxel.
\end{itemize}

\subsection{Operation state}\label{subsec:operation-state}
The operation state of the classifying machine can either be \textbf{active} as
in~\cite{schiebener2011segmentation,schiebener2012discovery} where robots can
move objects to find a segmentation or \textbf{passive}, where the received
image cannot be influenced. Among the passive algorithms, some segment in a
completely \textbf{automatic} fashion, others work in an \textbf{interactive}
mode. One example would be a system where the user clicks on the background or
marks a coarse segmentation and the algorithm finds a fine-grained
segmentation.
\cite{boykov2000interactive,rother2004grabcut,protiere2007interactive}~describe
systems which work in an interactive mode.

%!TEX root = vorlage.tex

\section{Evaluation and Datasets}\label{sec:evaluation-and-datasets}

%!TEX root = vorlage.tex

\subsection{Quality measures for evaluation}%
\label{subsec:quality-measures}%
A performance measure is a crucial part of any machine learning system. As
users of a semantic segmentation system expect correct results, the accuracy is
the most commonly used performance measure, but there are other measures of
quality which matter when segmentation algorithms are compared. This section
gives an overview of those quality measures.

\subsubsection{Accuracy}
\begin{figure}
\centering
\subfigure[Example Scene \label{fig:segmentation-example-scene}]{
  \includegraphics[width=0.45\linewidth, keepaspectratio]{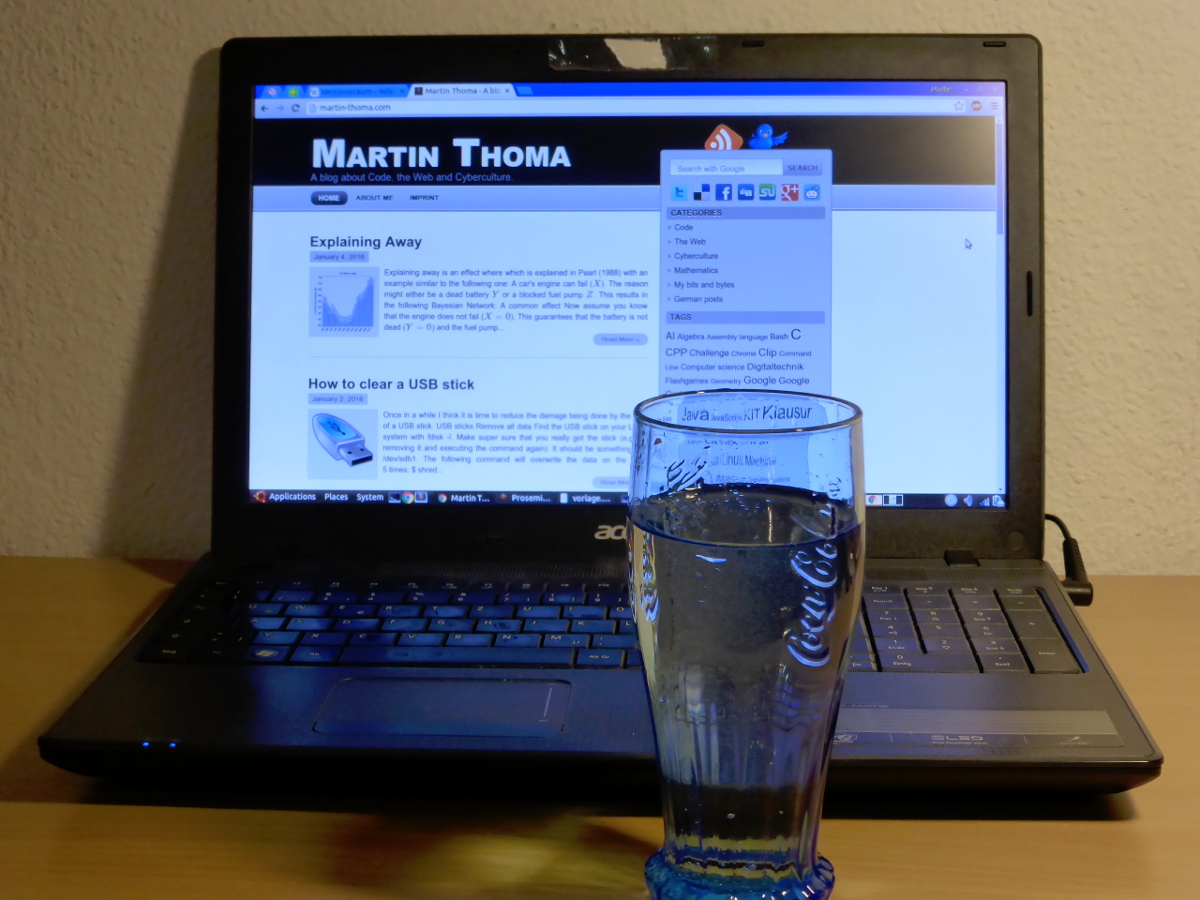}
}%
\subfigure[Visualization of a found segmentation \label{fig:segmentation-example-seg}]{
  \includegraphics[width=0.45\linewidth, keepaspectratio]{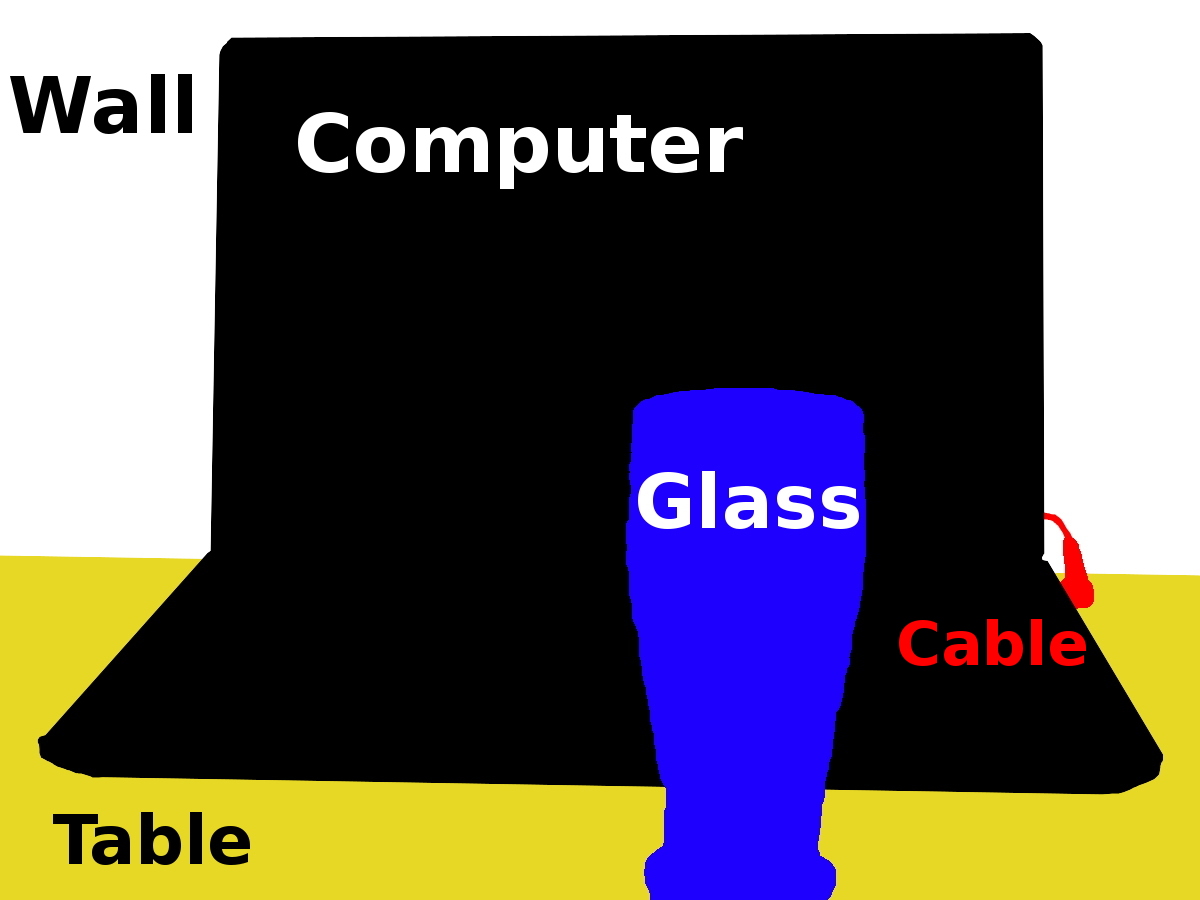}
}
\caption{An example of a scene and a possible visualization of a found segmentation.}
\label{fig:segmentation-example}
\end{figure}

Showing the correctness of the segmentation hypotheses is done in most
publications about semantic segmentation. However, there are a couple of
different ways how this accuracy can be displayed. One way to give readers a
first qualitative impression of the obtained segmentations is by showing
examples such as~\cref{fig:segmentation-example}.

However, this can only support the explanation of particular problems or
showcase special situation. For meaningful information about the overall
accuracy, there are a couple of metrics how accuracy can be defined.

For this section, let $k \in \mathbb{N}$ be the number of classes, $n_{ij} \in
\mathbb{N}_0$ with $i,j \in 1, \dots, k$ be the number of pixels which belong
to class~$i$ and were labeled as class~$j$. $(n_{ij})$ is called a
\textit{confusion matrix}. Let $t_i = \sum_{j=1}^k n_{ij}$ be the total number
of pixels of class~$i$.

One way to compare segmentation algorithms is by the pixel-wise accuracy of the
predicted segmentation as done in many publications
\cite{shotton2006textonboost,csurka2008simple,long2014fully}. This is also
called per-pixel rate and defined as $\frac{\sum_{i=1}^k n_{ii}}{\sum_{i=1}^k
t_i}$. Taking the pixel-wise classification accuracy has two major drawbacks:

\begin{problemnr}
    \item \label{item:problem-large-regions} Tasks like segmenting images for
          autonomous cars have large regions which have one class. This makes
          achieving classification accuracies of more than \SI{30}{\percent}
          with a priori knowledge only possible. For example, a system might
          learn that a certain position of the image is most of the time
          \enquote{sky} while another position is most of the time
          \enquote{road}.
    \item \label{item:problem-labeling-granularity} The manually labeled images
          could have a more coarse labeling. For example, a human classifier
          could have labeled a region as
          \enquote{car} and the algorithm could have split that region into
          the general \enquote{car} and the more specific \enquote{wheel of a
          car}
\end{problemnr}
\goodbreak
Three accuracy metrics which do not suffer from
\cref{item:problem-large-regions} are used in~\cite{long2014fully}:\nobreak%
\begin{itemize}
    \item \textit{mean accuracy}: $\frac{1}{k} \cdot \sum_{i=1}^k \frac{n_{ii}}{t_i} \in [0, 1]$
    \item \textit{mean intersection over union}: \hfill\\$\frac{1}{k} \cdot \sum_{i=1}^k \frac{n_{ii}}{t_i - n_{ii} + \sum_{j=1}^k n_{ji}} \in [0, 1]$
    \item \textit{frequency weighted intersection over union}:
          ${({\sum_{i=1}^k t_i})}^{-1} \sum_{i=1}^k t_i \cdot \frac{n_{ii}}{t_i - n_{ii} + \sum_{j=1}^k n_{ji}} \in [0, 1]$
\end{itemize}

Another problem might be pixels which cannot be assigned to one of the known
classes. For this reason, \cite{shotton2006textonboost} makes use of a void
class. This class gets completely ignored for all quality measures. Hence the
total number of pixels is assumed to be $\text{width} \cdot \text{height} - \text{number of void pixels}$.

One way to deal with \cref{item:problem-large-regions} and
\cref{item:problem-labeling-granularity} is giving the confusion matrix
as done in \cite{shotton2006textonboost}. However, this approach is not
feasible if many classes are given.

The $F$-measure is useful for binary classification task such as the KITTI road
segmentation benchmark~\cite{Fritsch2013ITSC} or crypt segmentation as done
by~\cite{cohen2015memory}. It is calculated as \enquote{the harmonic mean of
the precision and recall}~\cite{pantofaru2005comparison}:
\[F_\beta = (1+\beta)^2 \frac{\text{tp}}{(1+\beta^2)\cdot \text{tp}+ \beta^2 \cdot \text{fn} + \text{fp}}\]
where $\beta=1$ is chosen in most cases and \texttt{tp} means \textit{true
positive}, \texttt{fn} means \textit{false negative} and  \texttt{fp} means
\textit{false positive}.

Finally, it should be noted that a lot of other measures for the accuracy of
segmentations were proposed for non-semantic segmentation. One of those
accuracy measures is \textit{Normalized Probabilistic Rand}~(NPR) index which
was introduced in \cite{unnikrishnan2005measure} and evaluated
in~\cite{celebi2009improved} on dermoscopy images. Other non-semantic
segmentation measures were introduced in~\cite{martin2001database}, but the
reason for creating them seems to be to deal with the under-defined task
description of non-semantic segmentation. These accuracy measures try to deal
with different levels of coarsity of the segmentation. This is much less of a
problem in semantic segmentation and thus those measures are not explained
here.

\subsubsection{Speed}%
\label{subsubsec:speed-quality-measure}%
A maximum upper bound on the execution time for the inference on a single image
is a hard requirement for some applications. For example, in the case of
autonomous cars an algorithm which classifies pixel as street or no-street
and thus makes a semantic segmentation, every image needs to be processed
within \SI{20}{\milli\second}~\cite{bittel2015pixel}. This time is called
\textbf{latency}.

Most papers do not give exact values for the time their application needs. One
reason might be that this is very hardware, implementation and in some cases
even data specific. For example, \cite{hoover1996experimental} notes that their
algorithm needs \SI{10}{\second} on a Sun SparcStation~20. The fastest CPU ever
produced for this system had~\SI{200}{\mega\hertz}. Comparing this directly
with results which were obtained using an Intel~i7-4820K with
\SI{3.9}{\giga\hertz} would not be meaningful.

However, it does still make sense to mention the execution time as well as the
hardware in individual papers. This gives the interested reader the possibility
to estimate how difficult it might be to adjust the algorithm to work in the
required time-constraints.

Besides the latency, the \textbf{throughput} is another relevant characteristic
of algorithms and implementations for semantic segmentation. For example, for
the automatic description of images in order to enable text search the
throughput is of much higher importance than latency.

\subsubsection{Stability}%
\label{subsubsec:stability-quality-measure}%
A reasonable requirement on semantic segmentation algorithms is the stability
of a segmentation over slight changes in the input image. When the image data
is sightly blurred by smoke such as in~\cref{fig:smoke}, the segmentation
should not change. Also, two images which show a slight change in perspective
should also only result in slight changes in the
segmentation~\cite{pantofaru2005comparison}.

\subsubsection{Memory usage}
Peak memory usage matters when segmentation algorithms are used in devices like
smartphones or cameras, or when the algorithms have to finish in a given time
frame, run on the \gls{GPU} and consume so much memory for single image
segmentation that only the latest graphic cards can be used. However, no
publication were available mentioning the peak memory usage.

%!TEX root = vorlage.tex

\subsection{Datasets}

The computer vision community produced a couple of different datasets which are
publicly available. In the following, only the most widely used ones as well as
three medical databases are described. An overview over the quantity and
the kind of data is given by
\cref{table:segmentation-databases}.

\subsubsection{PASCAL VOC}

The PASCAL\footnote{\textbf{p}attern \textbf{a}nalysis, \textbf{s}tatistical
modelling and \textbf{c}omput\textbf{a}tional \textbf{l}earning, an EU network
of excellence} VOC\footnote{\textbf{V}isual \textbf{O}bject \textbf{C}lasses}
challenge was organized eight times with different datasets: Once every year
from 2005 to 2012~\cite{pascal-voc-2012}. Beginning with~2007, a segmentation
challenge was added~\cite{pascal-voc-2007}.

The dataset consists of annotated photographs from www.flicker.com, a photo
sharing website. There are multiple challenges for PASCAL VOC\@. The 2012
competition had five~challenges of which one is a segmentation challenge where
a single class label was given for each pixel. The classes are: aeroplane,
bicycle, bird, boat, bottle, bus, car, cat, chair, cow, dining table, dog,
horse, motorbike, person, potted plant, sheep, sofa, train, tv/monitor.

Although no new competitions will be held, new algorithms can be evaluated on
the 2010, 2011 and 2012 data via
\href{http://host.robots.ox.ac.uk:8080/}{http://host.robots.ox.ac.uk:8080/}

The PASCAL VOC segmentation challenges use the \textit{segmentation over union}
criterion (see \cref{subsec:quality-measures}).

\subsubsection{MSRCv2}\label{subsubsec:MSRCv2}

Microsoft Research has published a database of 591~photographs with pixel-level
annotation of 21~classes: aeroplane, bike, bird, boat, body, book, building,
car, cat, chair, cow, dog, face, flower, grass, road, sheep, sign, sky, tree,
water. Additionally, there is a \texttt{void} label for pixels which do not
belong to any of the 21~classes or which are close to the segmentation
boundary. This allows a \enquote{rough and quick hand-segmentation which does
not align exactly with the object boundaries}~\cite{shotton2006textonboost}.

\subsubsection{Medical Databases}

The Warwick-QU Dataset consists of 165~images with pixel-level annotation of
5~classes: \enquote{healthy, adenomatous, moderately differentiated,
moderately-to-poorly differentiated, and poorly
differentiated}~\cite{coelho2009nuclear}. This dataset is part of the
Gland Segmentation (GlaS) challenge.

The DIARETDB1~\cite{kalesnykiene2014diaretdb1} is a dataset of 89~images fundus
images. Those images show the interior surface of the eye. Fundus images can
be used to detect diabetic retinopathy. The images have four classes of coarse
annotations: hard and soft exudates, hemorrhages and red small dots.

20~test and additionally 20~training retinal fundus images are available
through the DRIVE data set~\cite{staal2004ridge}. The vessels were annotated.
Additionally, \cite{azzopardi2011detection} added vascular features.

The Open-CAS Endoscopic Datasets~\cite{maier2014can} are 60~images taken from
laparoscopic adrenalectomies and 60~images taken from laparoscopic pancreatic
resections. Those are from 3 surgical procedures each. Half of the data was
annotated by a medical expert for \enquote{medial instrument} and \enquote{no
medical instrument}. All images were labeled by anonymous untrained workers to
which they refer to as \textit{knowledge workers} (KWs). One crowd annotation
was obtained for each image by a majority vote on a pixel basis of
10~segmentations given by 10~different KWs.

%!TEX root = vorlage.tex

\section{Segmentation Pipeline}

\begin{figure}
    \centering
    \includegraphics[width=\linewidth, keepaspectratio]{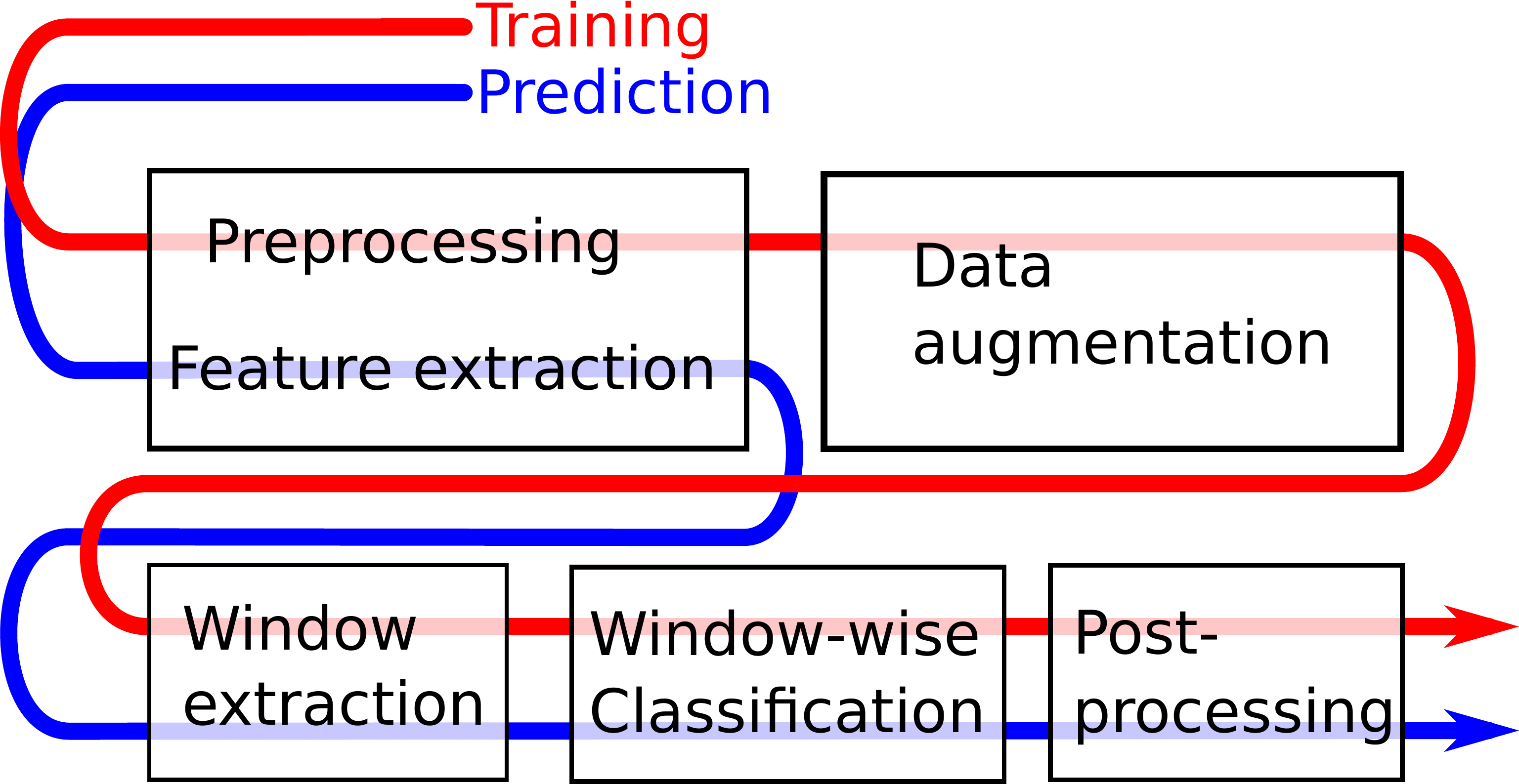}
    \caption{A typical segmentation pipeline gets raw pixel data, applies
             preprocessing techniques like scaling and feature extraction like
             HOG features. For training, data augmentation techniques such as
             image rotation can be applied. For every single image, patches
             of the image called \textit{windows} are extracted and those
             windows are classified. The resulting semantic segmentation can
             be refined by simple morphologic operations or by more complex
             approaches such as \glspl{MRF}.}
    \label{fig:segmentation-pipeline}
\end{figure}

Typically, semantic segmentation is done with a classifier which operates on
fixed-size feature inputs and a \textit{sliding-window}
approach~\cite{1467360,5490399,schroff2008object}. This means a classifier is
trained on images of a fixed size. The trained classifier is then fed with
rectangular regions of the image which are called \textit{windows}. Although
the classifier gets an image patch of e.g. $\SI{51}{\pixel} \times
\SI{51}{\pixel}$ of the environment, it might only classify the center pixel or
a subset of the complete window. This
segmentation pipeline is visualized in~\cref{fig:segmentation-pipeline}.

This approach was taken by~\cite{bittel2015pixel} and a majority of the VOC2007
participants~\cite{pascal-voc-2007}. As this approach has to apply the patch
classifier $512 \cdot 512 = \num{262144}$ times for images of size
$\SI{512}{\pixel} \times \SI{512}{\pixel}$, there are techniques for speeding
it up such as applying a stride and interpolating the results.

Neural networks are able to apply the sliding window approach in a very
efficient way by handling a trained network as a convolution and applying the
convolution on the complete image.

However, there are alternatives. Namely \glspl{MRF} and \glspl{CRF}
which take the information of the complete image and segment it in an holistic
approach.

%!TEX root = vorlage.tex

\section{Traditional Approaches}\label{sec:traditional-approaches}%
Image segmentation algorithms which use traditional approaches, hence don't
apply neural networks and make heavy use of domain knowledge, are wide-spread
in the computer vision community. Features which can be used for segmentation
are described in \cref{subsec:features}, a very brief overview of unsupervised,
non-semantic segmentation is given in
\cref{subsec:unsupervised-traditional-segmentation}, Random Decision
Forests are described in \cref{subsec:random-forests}, Markov Random Fields in \cref{subsec:markov-random-fields}
 and \glspl{SVM} in
\cref{subsec:trad-SVM}.
Postprocessing is covered in \cref{subsec:post-processing-methods}.

It should be noted that algorithms can use combination of methods. For example,
\cite{tighe2014scene} makes use of a combination of a \gls{SVM} and a
\gls{MRF}. Also, auto-encoders can be used to learn features which in turn
can be used by any classifier.

%!TEX root = vorlage.tex

\subsection{Features and Preprocessing methods}\label{subsec:features}%
The choice of features is very important in traditional approaches.
The most commonly used local and global features are explained in the following
as well as feature dimensionality reduction algorithms.

\subsubsection{Pixel Color}
Pixel color in different image spaces (\eg 3~features for RGB, 3~features for
HSV, 1~feature for the gray-value) are the most widely used features. A typical
image is in the RGB color space, but depending on the classifier and the
problem another color space might result in better segmentations. RGB, YcBcr,
HSL, Lab and YIQ are some examples used by \cite{cohen2015memory}. No single
color space has been proven to be superior to all others in all
contexts~\cite{cheng2001color}. However, the most common choices seem to be RGB
and HSI\@. Reasons for choosing RGB is simplicity and the support by
programming languages, whereas the choice of the HSI color space might make it
simpler for the classifier to become invariant to illumination. One reason for
choosing CIE-L*a*b* color space is that it approximates human perception of
brightness~\cite{kasson1992analysis}. It follows that choosing the L*a*b color
space helps algorithms to detect structures which are seen by humans. Another
way of improving the structure within an image is histogram equalization, which
can be applied to improve contrast~\cite{pizer1987adaptive,4228537}.

\subsubsection{Histogram of oriented Gradients}
\Gls{HOG} features interpret the image as a discrete function
$I: \mathbb{N}^2 \rightarrow \Set{0, \dots, 255}$ which maps the position $(x,
y)$ to a color. For each pixel, there are two gradients: The partial derivative
of $x$ and $y$. Now the original image is transformed to two feature maps of
equal size which represents the gradient. These feature maps are splitted into
patches and a histogram of the directions is calculated for each patch.
\gls{HOG} features were proposed in~\cite{1467360} and are used
in~\cite{bourdev2010detecting,felzenszwalb2010object} for segmentation tasks.

\subsubsection{SIFT}
\Gls{SIFT} feature descriptors describe keypoints in an image. The image patch
of the size $16 \times 16$ around the keypoint is taken. This patch is divided
in $16$ distinct parts of the size $4 \times 4$. For each of those parts a
histogram of 8~orientations is calculated similar as for \gls{HOG} features.
This results in a 128-dimensional feature vector for each keypoint.

It should be emphasized that SIFT is a global feature for a complete image.

\Gls{SIFT} is described in detail in~\cite{raey} and are used in~\cite{plath2009multi}.

\subsubsection{BOV}
\Gls{BOV}, also called \textit{bag of keypoints}, is based on vector
quantization. Similar to \gls{HOG} features, \gls{BOV} features are histograms
which count the number of occurrences of certain patterns within a patch of the
image. \Gls{BOV} are described in~\cite{csurka2004visual} and used in
combination with \gls{SIFT} feature descriptors in~\cite{csurka2008simple}.

\subsubsection{Poselets}
\textit{Poselets} rely on manually added extra keypoints such as \enquote{right
shoulder}, \enquote{left shoulder}, \enquote{right knee} and \enquote{left
knee}. They were originally used for human pose estimation. Finding those extra
keypoints is easily possible for well-known image classes like humans. However,
it is difficult for classes like airplanes, ships, organs or cells where the
human annotators do not know the keypoints. Additionally, the keypoints have to
be chosen for every single class. There are strategies to deal with those
problems like viewpoint-dependent keypoints. Poselets were used
in~\cite{bourdev2010detecting} to detect people and in~\cite{brox2011object}
for general object detection of the PASCAL VOC dataset.

\subsubsection{Textons}\label{subsubsec:textons}
A \textit{texton} is the minimal building block of vision. The computer vision
literature does not give a strict definition for textons, but edge detectors
could be one example. One might argue that deep learning techniques with
\glspl{CNN} learn textons in the first filters.

An excellent explanation of textons can be found in~\cite{zhu2005textons}.

\subsubsection{Dimensionality Reduction}
High-resolution images have a lot of pixels. Having one or more feature per
pixel results in well over a million features. This makes training difficult
while the higher resolution might not contain much more information. A simple
approach to deal with this is downsampling the high-resolution image to a
low-resolution variant. Another way of doing dimensionality reduction is
\gls{PCA}, which is applied by~\cite{chen2011pixel}. The idea behind \gls{PCA}
is to find a hyperplane on which all feature vectors can be projected with a
minimal loss of information. A detailed description of \gls{PCA} is given
by~\cite{smith2002tutorial}.

One problem of \gls{PCA} is the fact that it does not distinguish different
classes. This means it can happen that a perfectly linearly separable set of
feature vectors becomes not separable at all after applying \gls{PCA}.

There are many other techniques for dimensionality reduction. An overview and
a comparison over some of them is given by~\cite{van2009dimensionality}.

%!TEX root = vorlage.tex

\subsection{Unsupervised Segmentation}%
\label{subsec:unsupervised-traditional-segmentation}%

Unsupervised segmentation algorithms can be used in supervised segmentation as
another source of information or to refine a segmentation. While unsupervised
segmentation algorithms can never be semantic, they are well-studied and
deserve at least a very brief overview.

Semantic segmentation algorithms store information about the classes they were
trained to segment while non-semantic segmentation algorithms try to detect
consistent regions or region boundaries.

\subsubsection{Clustering Algorithms}
Clustering algorithms can directly be applied on the pixels, when one gives
a feature vector per pixel. Two clustering algorithms are $k$-means and the
mean-shift algorithm.

The $k$-means algorithm is a general-purpose clustering algorithm which
requires the number of clusters to be given beforehand. Initially, it places
the $k$ centroids randomly in the feature space. Then it assigns each
data point to the nearest centroid, moves the centroid to the center of the
cluster and continues the process until a stopping criterion is reached. A
faster variant is described in \cite{hartigan1975clustering}.

$k$-means was applied by~\cite{chen1998image} for medical image segmentation.

Another clustering algorithm is the mean-shift algorithm which was introduced
by~\cite{comaniciu2002mean} for segmentation tasks. The algorithm finds the
cluster centers by initializing centroids at random seed points and iteratively
shifting them to the mean coordinate within a certain range. Instead of taking
a hard range constraint, the mean can also be calculated by using any kernel.
This effectively applies a weight to the coordinates of the points. The mean
shift algorithm finds cluster centers at positions with a highest local
density of points.

\subsubsection{Graph Based Image Segmentation}%
\label{subsec:graph-based-image-segmentation}%
Graph-based image segmentation algorithms typically interpret pixels as
vertices and an edge weight is a measure of dissimilarity such as the
difference in color~\cite{felzenszwalb2004efficient,FelzenszwalbGraphCode}.
There are several different candidates for edges. The 4-neighborhood (north,
east, south west) or an 8-neighborhood (north, north-east, east, south-east,
south, south-west, west, north-west) are plausible choices.
One way to cut the edges is by building a minimum spanning tree and removing
edges above a threshold. This threshold can either be constant, adapted to the
graph or adjusted by the user. After the edge-cutting step, the connected
components are the segments.

A graph-based method which ranked \nth{2} in the Pascal VOC 2010
challenge~\cite{everingham2010pascal} is described
in~\cite{carreira2010constrained}. The system makes heavy use of the multi-cue
contour detector globalPb~\cite{4587420} and needs about \SI{10}{\giga\byte}
of main memory~\cite{Carreira2011}.

\subsubsection{Random Walks}

Random walks belong to the graph-based image segmentation algorithms. Random
walk image segmentation usually works as follows: Seed points are placed
on the image for the different objects in the image. From every single pixel,
the probability to reach the different seed points by a random walk is
calculated. This is done by taking image gradients as described in
\cref{subsec:features} for \gls{HOG} features. The class of the pixel is the
class of which a seed point will be reached with highest probability. At first,
this is an interactive segmentation method, but it can be extended to be
non-interactive by using another segmentation methods output as seed points.

\subsubsection{Active Contour Models}

\Glspl{ACM} are algorithms which segment images roughly along edges, but also
try to find a border which is smooth. This is done by defining a so called
\textit{energy function} which will be minimized. They were initially
described in~\cite{kass1988snakes}. \Glspl{ACM} can be used to segment an image
or to refine segmentation as it was done in~\cite{atkins1998fully} for brain
\gls{MR} images.

\subsubsection{Watershed Segmentation}\label{subsec:watershed}
The watershed algorithm takes a grayscale image and interprets it as a height
map. Low values are catchment basins and the higher values between two
neighboring catchment basins is the watershed. The catchment basins should
contain what the developer wants to capture. This implies that those areas
must be dark on grayscale images. The algorithm starts to fill the basins from
the lowest point. When two basins are connected, a watershed is found. The
algorithm stops when the highest point is reached.

A detailed description of the watershed segmentation algorithm is given
in~\cite{roerdink2000watershed}.

The watershed segmentation was used in~\cite{1260033} to segment white blood
cells. As the authors describe, the segmentation by watershed transform has
two flaws: Over-segmentation due to local minima and thick watersheds due to
plateaus.

%!TEX root = vorlage.tex

\subsection{Random Decision Forests}\label{subsec:random-forests}

Random Decision Forests were first proposed in~\cite{ho1995random}. This type
of classifier applies techniques called \textit{ensemble learning}, where
multiple classifiers are trained and a combination of their hypotheses is
used. One ensemble learning technique is the \textit{random subspaces} method
where each classifier is trained on a random subspace of the feature~space.
Another ensemble learning technique is \textit{bagging}, which is training the
trees on random subsets of the training~set. In the case of Random Decision
Forests, the classifiers are decision trees. A decision tree is a tree where
each inner node uses one or more features to decide in which branch to descend.
Each leaf is a class.

One strength of Random Decision Forests compared to many other classifiers like
\glspl{SVM} and neural networks is that the scale of measure of the features
(nominal, ordinal, interval, ratio) can be arbitrary. Another advantage of
Random Decision Forests compared to \glspl{SVM}, for example, is the speed
of training and classification.

Decision trees were extensively studied in the past 20~years and a
multitude of training algorithms have been proposed (e.g. ID3
in~\cite{quinlan1986induction}, C4.5 in~\cite{quinlan2014c4}). Possible
training hyperparameters are the measure to evaluate the \enquote{goodness of
split}~\cite{raey89empirical}, the number of decision trees being used, and if
the depth of the trees is restricted. Typically in the context of
classification, decision trees are trained by adding new nodes until
each leaf contains only nodes of a single class or until it is not possible to
split further. This is called a \textit{stopping criterion}.

There are two typical training modes: \textit{Central axis projection} and
\textit{perceptron training}. In training, for each node a hyperplane is
searched which is optimal according to an error function.

Random Decision Forests with texton features (see \cref{subsubsec:textons}) are
applied in~\cite{shotton2008semantic} for segmentation. In the~\cite{MSCR-db}
dataset, they report a per-pixel accuracy rate of \SI{66.9}{\percent} for their
best system. This system requires \SI{415}{\milli\second} for the segmentation of
$\SI{320}{\pixel} \times \SI{213}{\pixel}$ images on a single
\SI{2.7}{\giga\hertz} core. On the Pascal VOC~2007 dataset, they report an
average per-pixel accuracy for their best segmentation system
of~\SI{42}{\percent}.

An excellent introduction to Random Decision Forests for semantic segmentation
is given by~\cite{schroff2008object}.

%!TEX root = vorlage.tex

\subsection{SVMs}\label{subsec:trad-SVM}%

\Glspl{SVM} are well-studied binary classifiers which can be described by five
central ideas. For those ideas, the training data is represented as
$(\mathbf{x}_i, y_i)$ where $\mathbf{x}_i$ is the feature vector and $y_i \in
\Set{-1, 1}$ the binary label for training example $i \in \Set{1, \dots, m}$.

\begin{enumerate}
    \item If data is linearly separable, it can be separated by a hyperplane.
          There is one hyperplane which maximizes the distance to the next
          datapoints (\textit{support vectors}). This hyperplane should be
          taken:\\
          \begin{equation*}
          \begin{aligned}
              \minimize_{\mathbf{w}, b}\,&\frac{1}{2} \|\mathbf{w}\|^2\\
              \text{s.t. }& \forall_{i=1}^m y_i \cdot \underbrace{(\langle \mathbf{w}, \mathbf{x}_i\rangle + b)}_{\mathclap{\sgn \text{ applied to this gives the classification}}} \geq 1
          \end{aligned}
          \end{equation*}
    \item Even if the underlying process which generates the features for the
          two classes is linearly separable, noise can make the data not
          separable. The introduction of \textit{slack variables} to relax the
          requirement of linear separability solves this problem. The trade-off
          between accepting some errors and a more complex model is weighted by
          a parameter $C \in \mathbb{R}_0^+$. The bigger $C$, the more errors
          are accepted. The new optimization problem is:
          \begin{equation*}
          \begin{aligned}
              \minimize_{\mathbf{w}}\,&\frac{1}{2} \|\mathbf{w}\|^2 + C \cdot \sum_{i=1}^m \xi_i\\
              \text{s.t. }& \forall_{i=1}^m y_i \cdot (\langle \mathbf{w}, \mathbf{x}_i\rangle + b) \geq 1 - \xi_i
          \end{aligned}
          \end{equation*}

          Note that \(0 \le \xi_i \le 1\) means that the data point is within
          the margin, whereas \(\xi_i \ge 1\) means it is misclassified. An
          \gls{SVM} with $C > 0$ is also called a \textit{soft-margin \gls{SVM}}.
    \item The primal problem is to find the normal vector $\mathbf{w}$ and the
          bias $b$. The dual problem is to express $\mathbf{w}$ as a linear
          combination of the training data $\mathbf{x}_i$:
          \[\mathbf{w} = \sum_{i=1}^m \alpha_i y_i \mathbf{x}_i\]
          where $y_i \in \Set{-1, 1}$ represents the class of the training
          example and $\alpha_i$ are Lagrange multipliers. The usage of
          Lagrange multipliers is explained with some examples
          in~\cite{smithlagrange}. The usage of the Lagrange multipliers
          $\alpha_i$ changes the optimization problem depend on the
          $\alpha_i$ which are weights for the feature vectors. It turns
          out that most $\alpha_i$ will be zero. The non-zero weighted vectors
          are called \textit{support vectors}.

          The optimization problem is now, according to~\cite{burges1998tutorial}:
          \begin{equation*}
          \begin{aligned}
              \maximize_{\alpha_i}\,& \sum_{i=1}^m \alpha_i - \frac{1}{2} \sum_{i=1}^m \sum_{j=1}^m \alpha_i \alpha_j y_i y_j \langle \mathbf{x}_i, \mathbf{x}_j \rangle\\
              \text{s.t. } & \forall_{i=1}^m 0 \leq \alpha_i \leq C\\
              \text{s.t. } & \sum_{i=1}^m \alpha_i y_i = 0
          \end{aligned}
          \end{equation*}
    \item Not every dataset is linearly separable. This problem is approached
          by transforming the feature vectors $\mathbf{x}$ with a non-linear
          mapping $\Phi$ into a higher dimensional (probably
          $\infty$-dimensional) space. As the feature vectors $\mathbf{x}$
          are only used within scalar product
          $\langle \mathbf{x}_i, \mathbf{x}_j \rangle$, it is not necessary to
          do the transformation. It is enough to do the calculation
          \[K(\mathbf{x}_i, \mathbf{x}_j) = \langle \mathbf{x}_i, \mathbf{x}_j \rangle\]

          This function $K$ is called a \textit{kernel}. The idea of never
          explicitly transforming the vectors $\mathbf{x}_i$ to the higher
          dimensional space is called the \textit{kernel trick}. Common kernels
          include the polynomial kernel
          \[K_P(\mathbf{x}_i, \mathbf{x}_j) = (\langle \mathbf{x}_i, \mathbf{x}_j \rangle + r)^p\]
          of degree $p$ and coefficient $r$, the Gaussian \gls{RBF} kernel
          \[K_{\text{Gauss}}(\mathbf{x}_i, \mathbf{x}_j) = e^{\frac{-\gamma\|\mathbf{x}_i - \mathbf{x}_j\|^2}{2 \sigma^2}}\]
          and the sigmoid kernel
          \[K_{\text{tanh}}(\mathbf{x}_i, \mathbf{x}_j) = \tanh(\gamma \langle \mathbf{x}_i, \mathbf{x}_j \rangle - r)\]
          where the parameter $\gamma$ determines how much influence single
          training examples have.
    \item The described \glspl{SVM} can only distinguish between two classes.
          Common strategies to expand those binary classifiers to multi-class
          classification is the \textit{one-vs-all} and the \textit{one-vs-one}
          strategy. In the one-vs-all strategy $n$ classifiers have to be
          trained which can distinguish one of the $n$ classes against all
          other classes. In the one-vs-one strategy $\frac{n^2 - n}{2}$
          classifiers are trained; one classifier for each pair of classes.
\end{enumerate}

A detailed description of \glspl{SVM} can be found in~\cite{burges1998tutorial}.

\Glspl{SVM} are used by \cite{yang2012layered} on the 2009 and 2010 PASCAL
segmentation challenge~\cite{everingham2010pascal}. They did not hand their
classifier in to the challenge itself, but calculated an average rank~of~7
among the different categories.

\cite{felzenszwalb2010object} also used an \gls{SVM} based method with \gls{HOG}
features and achieved the \nth{7}~rank in the 2010 PASCAL segmentation
challenge by mean accuracy. It needs about \SI{2}{\second} on a
\SI{2.8}{\giga\hertz} 8-core Intel processor.

%!TEX root = vorlage.tex

\subsection{Markov Random Fields}\label{subsec:markov-random-fields}
\tikzstyle{pixel}=[draw,black,circle,minimum size=10pt,inner sep=0pt,fill=red!50]
\tikzstyle{label}=[draw,black,circle,minimum size=10pt,inner sep=0pt,fill=blue!50]
\tikzstyle{edge}=[very thick]
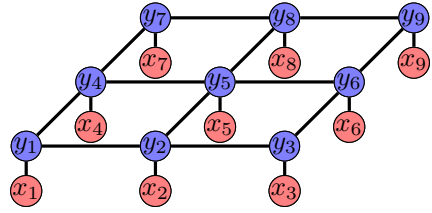
\begin{figure}
    \centering
\begin{tikzpicture}[scale=1.7]
    \node (x1)[pixel] at (0.0,1.15) {$x_1$};
    \node (x2)[pixel] at (1.0,1.15) {$x_2$};
    \node (x3)[pixel] at (2.0,1.15) {$x_3$};
    \node (x4)[pixel] at (0.5,1.65) {$x_4$};
    \node (x5)[pixel] at (1.5,1.65) {$x_5$};
    \node (x6)[pixel] at (2.5,1.65) {$x_6$};
    \node (x7)[pixel] at (1.0,2.15) {$x_7$};
    \node (x8)[pixel] at (2.0,2.15) {$x_8$};
    \node (x9)[pixel] at (3.0,2.15) {$x_9$};

    \node (y1)[label] at (0.0,1.5) {$y_1$};
    \node (y2)[label] at (1.0,1.5) {$y_2$};
    \node (y3)[label] at (2.0,1.5) {$y_3$};
    \node (y4)[label] at (0.5,2.0) {$y_4$};
    \node (y5)[label] at (1.5,2.0) {$y_5$};
    \node (y6)[label] at (2.5,2.0) {$y_6$};
    \node (y7)[label] at (1.0,2.5) {$y_7$};
    \node (y8)[label] at (2.0,2.5) {$y_8$};
    \node (y9)[label] at (3.0,2.5) {$y_9$};

    \draw[edge] (y1) -- (y2);
    \draw[edge] (y1) -- (y4);
    \draw[edge] (y2) -- (y3);
    \draw[edge] (y2) -- (y5);
    \draw[edge] (y3) -- (y6);
    \draw[edge] (y4) -- (y5);
    \draw[edge] (y4) -- (y7);
    \draw[edge] (y5) -- (y6);
    \draw[edge] (y5) -- (y8);
    \draw[edge] (y6) -- (y9);
    \draw[edge] (y7) -- (y8);
    \draw[edge] (y8) -- (y9);

    \draw[edge] (x1) -- (y1);
    \draw[edge] (x2) -- (y2);
    \draw[edge] (x3) -- (y3);
    \draw[edge] (x4) -- (y4);
    \draw[edge] (x5) -- (y5);
    \draw[edge] (x6) -- (y6);
    \draw[edge] (x7) -- (y7);
    \draw[edge] (x8) -- (y8);
    \draw[edge] (x9) -- (y9);

    %\draw [dashed] (-0.5,-0.3) -- (2,-0.3) -- (3.5,1.5) -- (0.5,1.5) -- (-0.5,-0.3);
    \node (x1)[pixel] at (0.0,1.15) {$x_1$};
    \node (x2)[pixel] at (1.0,1.15) {$x_2$};
    \node (x3)[pixel] at (2.0,1.15) {$x_3$};
    \node (x4)[pixel] at (0.5,1.65) {$x_4$};
    \node (x5)[pixel] at (1.5,1.65) {$x_5$};
    \node (x6)[pixel] at (2.5,1.65) {$x_6$};
    \node (x7)[pixel] at (1.0,2.15) {$x_7$};
    \node (x8)[pixel] at (2.0,2.15) {$x_8$};
    \node (x9)[pixel] at (3.0,2.15) {$x_9$};

    \node (y1)[label] at (0.0,1.5) {$y_1$};
    \node (y2)[label] at (1.0,1.5) {$y_2$};
    \node (y3)[label] at (2.0,1.5) {$y_3$};
    \node (y4)[label] at (0.5,2.0) {$y_4$};
    \node (y5)[label] at (1.5,2.0) {$y_5$};
    \node (y6)[label] at (2.5,2.0) {$y_6$};
    \node (y7)[label] at (1.0,2.5) {$y_7$};
    \node (y8)[label] at (2.0,2.5) {$y_8$};
    \node (y9)[label] at (3.0,2.5) {$y_9$};
\end{tikzpicture}
\caption{\gls{CRF} with 4-neighborhood. Each node $x_i$ represents a pixel and
         each node $y_i$ represents a label.}
\label{fig:crf-image}
\end{figure}
\Glspl{MRF} are undirected probabilistic graphical models which are wide-spread
model in computer vision. The overall idea of \glspl{MRF} is to assign a random
variable for each feature and a random variable for each pixel which gets
labeled as shown in~\cref{fig:crf-image}. For example, a \gls{MRF} which is
trained on images of the size
$\SI{224}{\pixel} \times \SI{224}{pixel}$ and gets the raw RGB values as
features has
\[\underbrace{224 \cdot 224 \cdot 3}_{\text{input}} + \underbrace{224 \cdot 224}_{\text{output}} = \num{200704}\]
random variables. Those random variables are conditionally independent, given
their local neighborhood. These (in)dependencies can be expressed with a graph.

Let $G=(\mathcal{V}, \mathcal{E})$ be the associated undirected graph of an
\gls{MRF} and $\mathcal{C}$ be the set of all maximal cliques in that graph.
Nodes represent random variables $\mathbf{x}, \mathbf{y}$ and edges represent
conditional dependencies. Just like in
\crefname{subsec:graph-based-image-segmentation}, the
4-neighborhood~\cite{shotton2006textonboost} and the 8-neighborhood are
reasonable choices for constructing the graph.

Typically, random variables $\mathbf{y}$ represent the class of a single pixel,
random variables $\mathbf{x}$ represent a pixel values and edges represent
pixel neighborhood in computer vision problems segmentation problems where
\glspl{MRF} are used. Accordingly, the random variables $\mathbf{y}$ live on
$1, \dots, \text{nr of classes}$ and the random variables $\mathbf{x}$
typically live on $0, \dots, 255$ or $[0, 1]$.

The probability of $\mathbf{x}, \mathbf{y}$ can be expressed as
\[P(\mathbf{x}, \mathbf{y}) = \frac{1}{Z} e^{-E(\mathbf{x}, \mathbf{y})}\]
where $Z = \sum_{\mathbf{x}, \mathbf{y}} e^{-E(\mathbf{x}, \mathbf{y})}$ is a normalization term called
the \textit{partition function} and $E$ is called the \textit{energy function}.
A common choice for the energy function is
\[E(\mathbf{x}, \mathbf{y}) = \sum_{c \in \mathcal{C}} \psi_c(\mathbf{x}, \mathbf{y})\]
where $\psi$ is called a \textit{clique potential}. One choice for cliques
of size two $\mathbf{x}, \mathbf{y} = (x_1, x_2)$ is~\cite{kato2006markov}
\[\psi_c(x_1, x_2) = w \delta(x_1, x_2) = \begin{cases}+w &\text{if } x_1 \neq x_2\\-w &\text{if } x_1 = x_2\end{cases}\]

According to~\cite{murphy2012machine}, the most common way of inference over
the posterior \gls{MRF} in computer vision problems is \gls{MAP} estimation.

Detailed introductions to \glspl{MRF} are given by
\cite{blake2011markov,murphy2012machine}. \glspl{MRF} are used by \cite{zhang2001segmentation} and \cite{moser2012markov}
for image segmentation.

% Characterizations of MRF:
% Label space: binary vs multi-label; homogeneous vs heterogeneous
% Order: unary vs pairwise vs higher-order
% Structure: chain vs tree vs grid vs general graph; neighborhood size
% Potentials: submodular, convex, compressible

% Markov Random Fields for Computer Vision (Part 1)
% http://users.cecs.anu.edu.au/~sgould/papers/part1-MLSS-2011.pdf  --- very nice!
% http://users.cecs.anu.edu.au/~sgould/papers/part2-MLSS-2011.pdf
% http://users.cecs.anu.edu.au/~sgould/papers/part3-MLSS-2011.pdf

% Markov Random Field Image Models and Their Applications to Computer Vision
% http://www.mathunion.org/ICM/ICM1986.2/Main/icm1986.2.1496.1517.ocr.pdf

\subsection{Conditional Random Fields}\label{subsec:conditional-random-fields}

\Glspl{CRF} are \glspl{MRF} where all clique potentials are conditioned on
input features~\cite{murphy2012machine}. This means, instead of learning the
distribution $P(\mathbf{y}, \mathbf{x})$, the task is reformulated to learn the
distribution $P(\mathbf{y}| \mathbf{x})$. One consequence of this reformulation
is that \glspl{CRF} need much less parameters as the distribution of
$\mathbf{x}$ does not have to be estimated. Another advantage of \glspl{CRF}
compared to \glspl{MRF} is that no distribution assumption about $\mathbf{x}$
has to be made.

A \gls{CRF} has the partition function $Z$:
\[Z(\mathbf{x}) = \sum_{\mathbf{y}} P(\mathbf{x}, \mathbf{y})\]

and joint probability distribution

\[P(\mathbf{y} | \mathbf{x}) = \frac{1}{Z(\mathbf{x})} \prod_{c \in \mathcal{C}} \psi_c(\mathbf{y}_c | \mathbf{x})\]

The simplest way to define the clique potentials $\psi$ is the count of the
class $\mathbf{y}_c$ given $\mathbf{x}$ added with a positive smoothing
constant to prevent the complete term from getting zero.

\Glspl{CRF} as described in~\cite{associative09} have reached top performance
in PASCAL VOC 2010~\cite{VOC2010Results} and are also used in
\cite{multiscale04,shotton2006textonboost} for semantic segmentation.

A method similar to \glspl{CRF} was proposed in~\cite{gonfaus2010harmony}.
The system of Gonfaus~et.al. ranked~\nth{1} by mean accuracy in the segmentation
task of the PASCAL VOC 2010 challenge~\cite{everingham2010pascal}.

An introduction to \glspl{CRF} is given by~\cite{sutton2011introduction}.

%!TEX root = vorlage.tex

\subsection{Post-processing methods}\label{subsec:post-processing-methods}%
Post-processing refine a found segmentation and remove obvious
errors. For example, the morphological operations \textit{opening} and
\textit{closing} can remove noise. The opening operation is a dilation followed
by a erosion. This removes tiny segments. The closing operation is a erosion
followed by a dilation. This removes tiny gaps in otherwise filled regions.
They were used in~\cite{chen1998image} for biomedical image segmentation.

Another way of refinement of the found segmentation is by adjusting the
segmentation to match close edges. This was used in~\cite{brox2011object} with
an ultra-metric contour map~\cite{arbelaez2009contours}.

Active contour models are another example of a post-processing
method~\cite{kass1988snakes}.

%!TEX root = vorlage.tex

\section{Neural Networks for Semantic Segmentation}\label{sec:nn}

Artificial neural networks are classifiers which are inspired by biologic
neurons. Every single artificial neuron has some inputs which are weighted and
sumed up. Then, the neuron applies a so called \textit{activation function} to
the weighted sum and gives an output. Those neurons can take either a feature
vector as input or the output of other neurons. In this way, they build up
feature hierarchies.

The parameters they learn are the \textit{weights} $w \in \mathbb{R}$. They are
learned by gradient descent. To do so, an error function --- usually
cross-entropy or mean squared error --- is necessary. For the gradient descent
algorithm, one sees the labeled training data as given, the weights as
variables and the error function as a surface in this weight-space. Minimizing
the error function in the weight space adapts the neural network to the
problem.

There are lots of ideas around neural networks like regularization, better
optimization algorithms, automatically building up architectures, design
choices for activation functions. This is not explained in detail here, but
some of the mayor breakthroughs are outlined.

\Glspl{CNN} are neural networks which learn image filters. They drastically
reduce the number of parameters which have to be learned while being still
general enough for the problem domain of images. This was shown by Alex
Krizhevsky~et~al. in~\cite{krizhevsky2012imagenet}. One major idea was a clever
regularization called \textit{dropout training}, which set the output of
neurons while training randomly to zero. Another contribution was the usage of
an activation function called \textit{rectified linear unit}:
\[\varphi_{\text{ReLU}}(x) = \max(0, x)\]
Those are much faster to train than the commonly used sigmoid activation
functions
\[\varphi_{\text{Sigmoid}}(x) = \frac{1}{e^{-x} + 1}\]
Krizhevsky~et~al. implemented those ideas and participated in the
\gls{ILSVRC}. The best other system, which used SIFT features and Fisher
Vectors, had a performance of about \SI{25.7}{\percent} while the network by
Alex Krizhevsky~et~al. got \SI{17.0}{\percent} error rate on the ILSVRC-2010
dataset. As a preprocessing step, they downsampled all images to a fixed size
of $\SI{256}{\pixel} \times \SI{256}{\pixel}$ before they fed the features into
their network. This network is commonly known as \textit{AlexNet}.

Since AlexNet was developed, a lot of different neural networks have been
proposed. One interesting example is~\cite{pinheiro2013recurrent}, where
a recurrent \gls{CNN} for semantic segmentation is presented.

Another notable paper is~\cite{long2014fully}. The algorithm presented there
makes use of a classifying network such as AlexNet, but applies the complete
network as an image filter. This way, each pixel gets a probability
distribution for each of the trained classes. By taking the most likely class,
a semantic segmentation can be done with arbitrary image sizes.

A very recent publication by Dai~et~al.~\cite{dai2015instance} showed that
segmentation with much deeper networks is possible and achieves better results.

More detailed explanations to neural networks for visual recognition is given
by \cite{CS231n}.

%!TEX root = vorlage.tex

\section{Possible Problems in the Data for Segmentation algorithms}%
\label{sec:problems}

Different segmentation workflows have different problems. However, there are
a couple of special cases which should be tested. Those cases might not occur
often in the training data, but it could still happen in the productive system.

I am not aware of any systematic work which examined the influence of problems
such as the following.

\subsection{Lens Flare}
\begin{figure}
\centering
\subfigure[Lens Flare\newline Image by \cite{image:wikipedia:lens-flare} \label{fig:lens-flare}]{
  \includegraphics[width=0.45\linewidth, keepaspectratio]{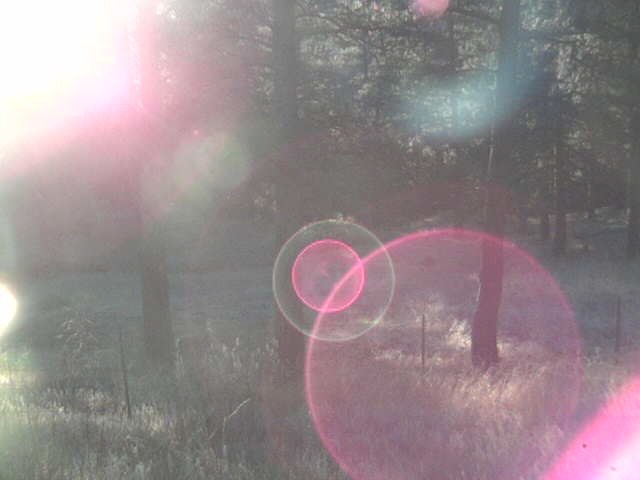}
}%
\subfigure[Vignetting\newline Image by \cite{image:wikipedia:vignetting} \label{fig:Vignetting}]{
  \includegraphics[width=0.45\linewidth, keepaspectratio]{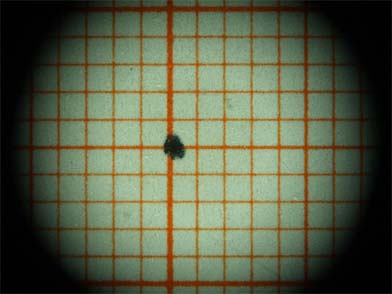}
}
\subfigure[Smoke by cauterization\newline Image by \cite{giannarou2013probabilistic} \label{fig:smoke}]{
  \includegraphics[width=0.45\linewidth, keepaspectratio]{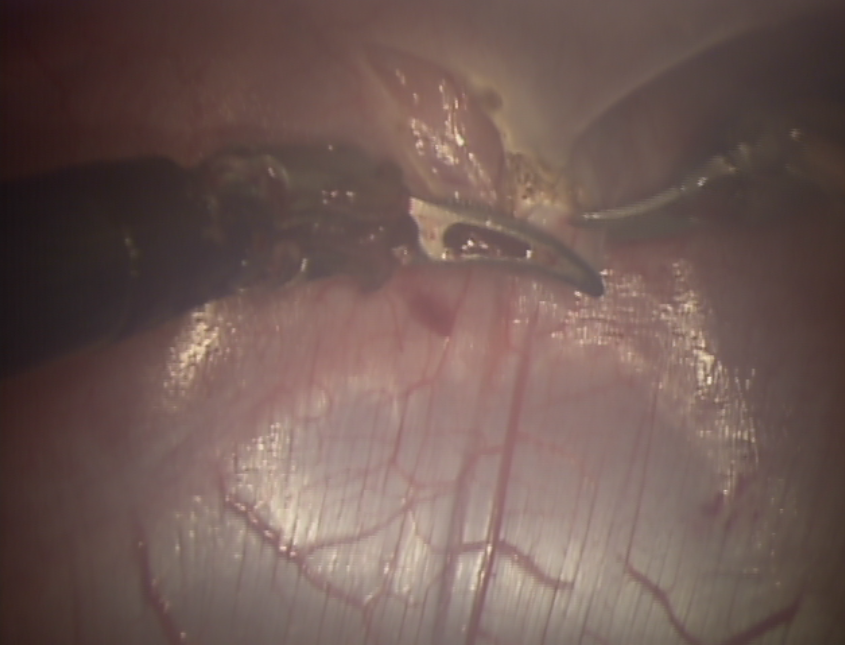}
}%
\subfigure[Camouflage\newline Image by \cite{image:wikipedia:camouflage} \label{fig:camouflage}]{
  \includegraphics[width=0.45\linewidth, keepaspectratio]{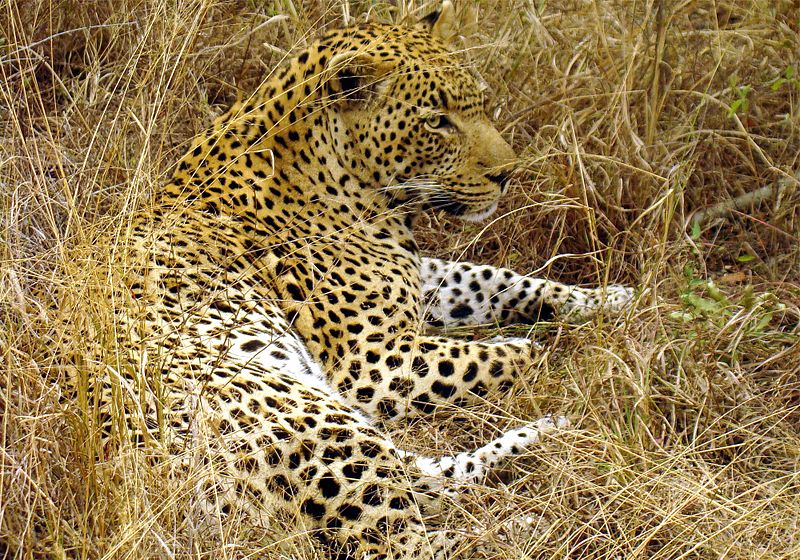}
}
\subfigure[Transparency \label{fig:transparency-glass}]{
  \includegraphics[width=0.45\linewidth, keepaspectratio]{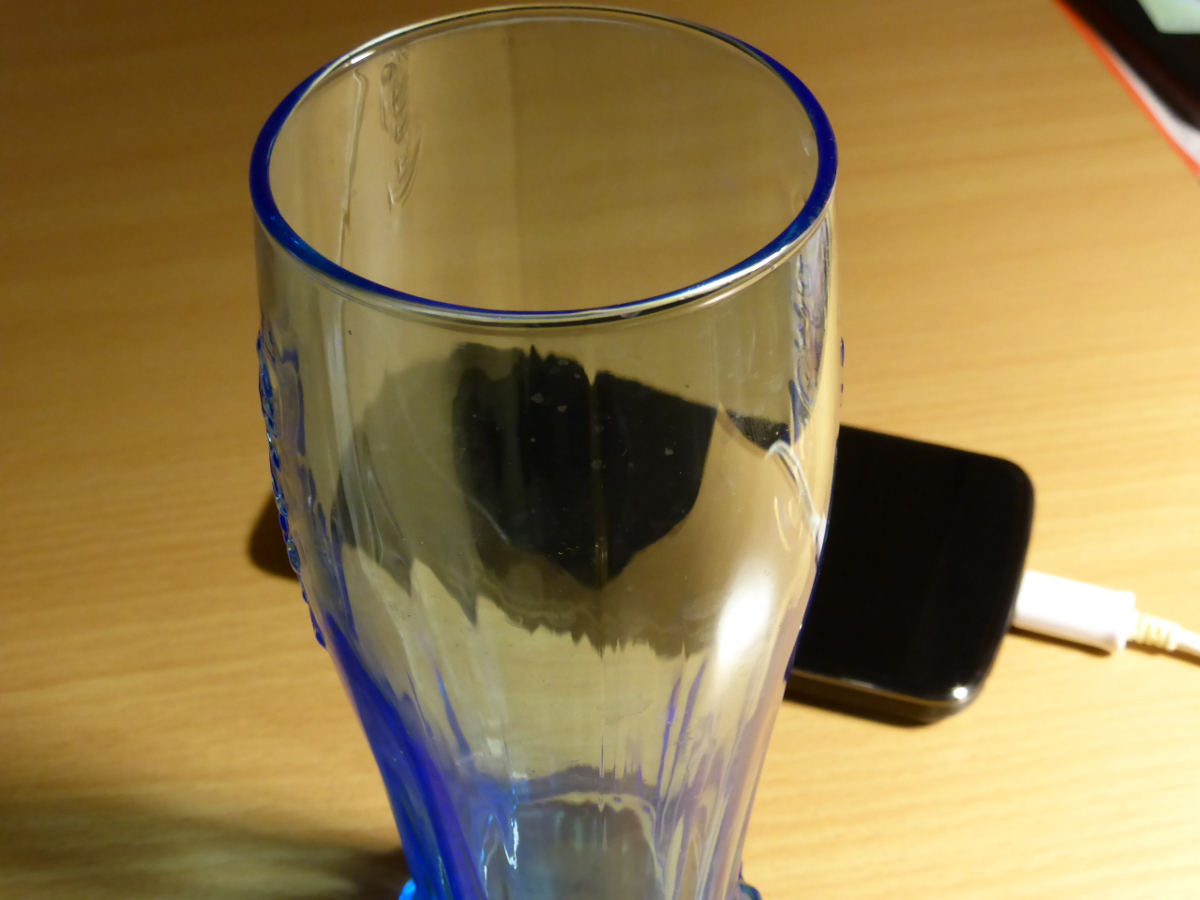}
}%
\subfigure[Viewpoint \label{fig:viewpoint}]{
  \includegraphics[width=0.45\linewidth, keepaspectratio]{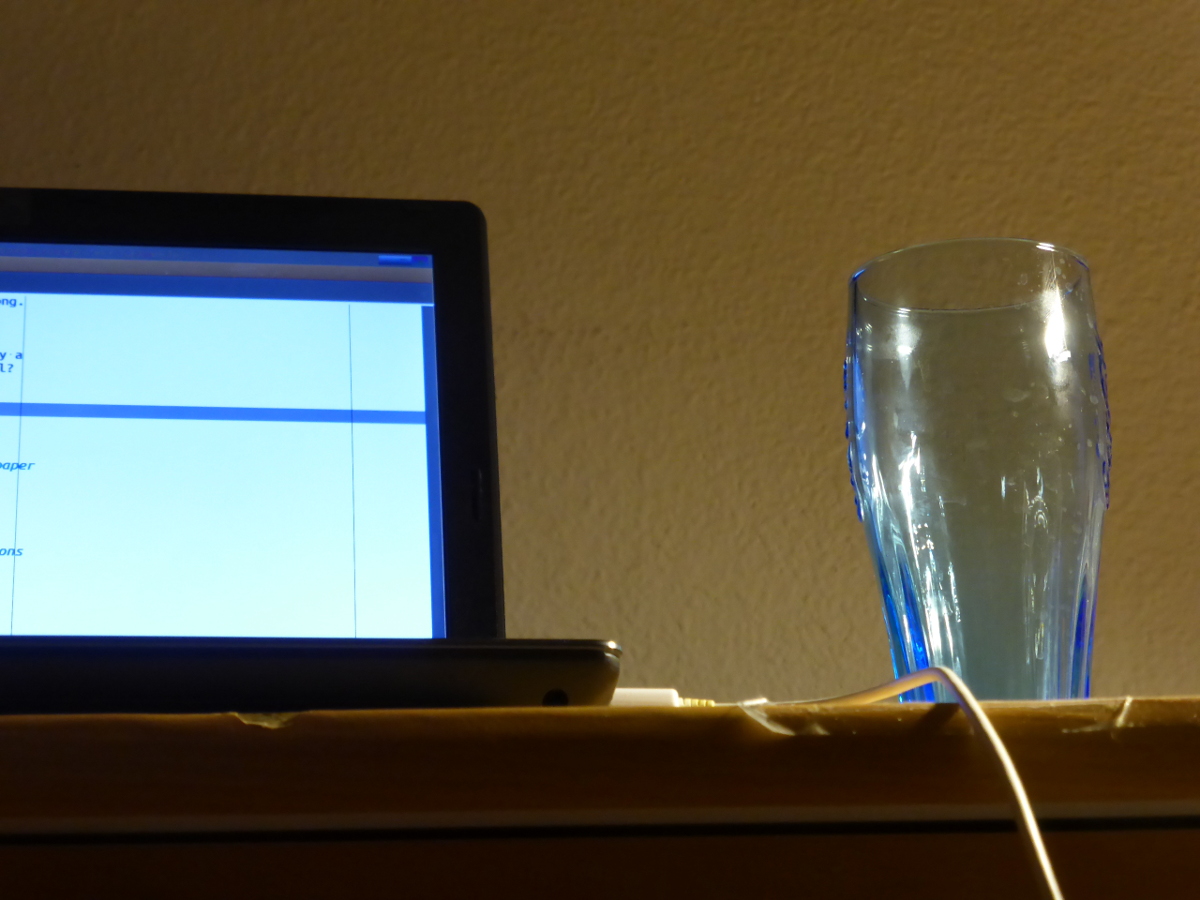}
}
\caption{Examples of images which might cause semantic segmentation systems to fail.}
\label{fig:test}
\end{figure}

Lens flare is the effect of light getting scattered in the lens system of the
camera. The testing data set of the KITTI road evaluation
benchmark~\cite{Fritsch2013ITSC} has a couple of photos with this problem.
\Cref{fig:lens-flare} shows an extreme example of lens flare.

\subsection{Vignetting}
Vignetting is the effect of a photograph getting darker in the corners. This
can have many reasons, for example filters on the camera blocking light at the
corners.

\subsection{Blurred images}
Images can be blurred for a couple of reasons. A problem with the lenses
mechanics, focusing on the wrong point, too quick movement, smoke or foam.
One example of a blurred image is \cref{fig:smoke}, which was taken during an
in~vivo porcine procedure of diaphragm dissection. The smoke was caused by
cauterization.

\subsection{Other Problems}
If the following effects can occur at all and if they are problems depends
heavily on the problem domain and the used model.

\subsubsection{Partial Occlusions}
Segmentation systems which employ a model of the objects which should be
segmented might suffer from partial occlusions.

\subsubsection{Camouflage}
Some objects, like animals in the wild, actively try to hide
(see~\cref{fig:camouflage} as an example). In other cases it might just be bad
luck that objects are hard for humans to detect. This problem has two
interesting aspects: On the one hand, the segmenting system might suffer from
the same problems as humans do. On the other hand, the segmenting system might
be better than humans are, but it is forced to learn from images labeled by
humans. If the labels are wrong, the system is forced to learn something wrong.

\subsubsection{Semi-transparent Occlusion}
Some objects like drinking glasses can be visible and still leave the object
behind them visible as shown in \cref{fig:transparency-glass}. This is mainly a
definition problem: Is the seen pixel the glass label or the smartphone label?

\subsubsection{Viewpoints}
Changes in viewpoints can be a problem, if they don't occur in the training
data. For example, an image captioning system which was trained on photographs
of professional photographers might not have photos from the point of view of
a child. This is visualized in \cref{fig:viewpoint}.

%!TEX root = vorlage.tex

\section{Discussion}%
\label{sec:discussion}
Ohta et al. wrote~\cite{ohta1978analysis} 38~years ago. It is one of the first
papers mentioning semantic segmentation. In this time, a lot of work was done
and many different directions have been explored. Different kinds of semantic
segmentation have emerged.

This paper presents a taxonomy of those kinds of semantic segmentation and a
brief overview of completely automatic, passive, semantic segmentation
algorithms.

Future work includes a comparative study of those algorithms on publicly
available dataset such as the ones presented
in~\cref{table:segmentation-databases}. Another open question is the influence
of the problems described in~\cref{sec:problems}. This could be done using a
subset of the thousands of images of Wikipedia Commons, such as \href{https://commons.wikimedia.org/wiki/Category:Blurring}{https://commons.wikimedia.org/wiki/Category:Blurring} for blurred images.

A combination of different classifiers in an ensemble would be an interesting
option to explore in order to improve accuracy. Another direction which is
currently studied is combining classifiers such as neural networks with
\glspl{CRF}~\cite{zheng2015conditional}.

\newpage
\bibliography{literature}
\bibliographystyle{IEEEtranSA}\vfill
% \columnbreak
\printglossaries%

%!TEX root = vorlage.tex

\clearpage\onecolumn
\begin{appendices}
\section{Tables}
\begin{table}[ht]
    \centering
    \begin{tabular}{lcrrcl}
    \toprule
    Database        & Image Resolution (width $\times$ height) & \parbox{1cm}{\centering Number of\\Images}  & \parbox{1cm}{\centering Number of\\Classes}  & Channels & Data source\\\midrule
    Colon Crypt DB  & $\hphantom{0}(\SIrange{302}{1116}{\pixel}) \times (\SIrange{349}{875}{\pixel})$             &  \num{389} &  2 & 3        & \cite{colon-crypt-segmentation-db}\\
    DIARETDB1       & $\hphantom{00}\SI{1500}{\pixel} \times \SI{1500}{\pixel}$                                   &   \num{89} &  4 & 3        & \cite{kalesnykiene2014diaretdb1}\\
    KITTI Road      & $(\SIrange{1226}{1242}{\pixel}) \times (\SIrange{370}{376}{\pixel})$                        &  \num{289} &  2 & 3        & \cite{Fritsch2013ITSC}\\
    MSRCv1          & $\hphantom{00}(\SIrange{213}{320}{\pixel}) \times (\SIrange{213}{320}{\pixel})$             &  \num{240} &  9 & 3        & \cite{MSRC-data}\\
    MSRCv2          & $\hphantom{00}(\SIrange{213}{320}{\pixel}) \times (\SIrange{162}{320}{\pixel})$             &  \num{591} & 23 & 3        & \cite{MSRC-data}\\
    Open-CAS Endoscopic Datasets & $\hphantom{00}\SI{640}{\pixel} \times \SI{480}{\pixel}$                        &  \num{120} &  2 & 3        & \cite{maier2014can}\\
    PASCAL VOC 2012 & $\hphantom{00}(\SIrange{142}{500}{\pixel}) \times (\hphantom{0}\SIrange{71}{500}{\pixel})$  & \num{2913} & 20 & 3        & \cite{pascal-voc-2012-data}  \\
    Warwick-QU      & $\hphantom{00}(\SIrange{567}{775}{\pixel}) \times (\SIrange{430}{522}{\pixel})$             &  \num{165} &  5 & 3        & \cite{coelho2009nuclear}\\
    \bottomrule
    \end{tabular}
    \caption{An overview over publicly available image databases with a semantic segmentation ground trouth.}
    \label{table:segmentation-databases}
\end{table}
\end{appendices}

\end{document}